\documentclass[10pt,twocolumn,letterpaper]{article}

\usepackage{iccv}
\usepackage{times}
\usepackage{epsfig}
\usepackage{graphicx}
\usepackage{amsmath}
\usepackage{amssymb}
\usepackage[misc]{ifsym}
\usepackage{multirow}

\usepackage[pagebackref=true,breaklinks=true,letterpaper=true,colorlinks,bookmarks=false]{hyperref}

\usepackage[accsupp]{axessibility}  
\usepackage{booktabs}
\usepackage{multirow}
\usepackage{algorithm}
\usepackage{algorithmic}

\iccvfinalcopy 

\def\iccvPaperID{10747} 

\ificcvfinal\fi

\begin{document}

\title{Dual Meta-Learning with Longitudinally Generalized Regularization for One-Shot Brain Tissue Segmentation Across the Human Lifespan}

\author{Yongheng Sun$^1$, Fan Wang$^2$\textsuperscript{(\Letter)}, Jun Shu$^1$, Haifeng Wang$^1$, Li Wang$^3$\textsuperscript{(\Letter)}, Deyu Meng$^1$, Chunfeng Lian$^1$\textsuperscript{(\Letter)}\\
$^1$School of Mathematics and Statistics, Xi'an Jiaotong University, Xi'an 710049, China\\
$^2$The Key Laboratory of Biomedical Information Engineering of Ministry of Education, \\School of Life Science and Technology, Xi'an Jiaotong University, Xi'an 710049, China\\
$^3$UNC Chapel Hill, Chapel Hill, NC 27599, United States\\
{\tt\small chunfeng.lian@xjtu.edu.cn, li\_wang@med.unc.edu, fan.wang@xjtu.edu.cn}
}


\maketitle
\ificcvfinal\thispagestyle{empty}\fi

\begin{abstract}
   Brain tissue segmentation is essential for neuroscience and clinical studies. However, segmentation on longitudinal data is challenging due to dynamic brain changes across the lifespan. Previous researches mainly focus on self-supervision with regularizations and will lose longitudinal generalization when fine-tuning on a specific age group. In this paper, we propose a dual meta-learning paradigm to learn longitudinally consistent representations and persist when fine-tuning. Specifically, we learn a plug-and-play feature extractor to extract longitudinal-consistent anatomical representations by meta-feature learning and a well-initialized task head for fine-tuning by meta-initialization learning. Besides, two class-aware regularizations are proposed to encourage longitudinal consistency.
   Experimental results on the iSeg2019 and ADNI datasets demonstrate the effectiveness of our method. Our code is available at \url{https://github.com/ladderlab-xjtu/DuMeta}.

\end{abstract}

\section{Introduction}
\begin{figure}[t]
\setlength{\abovecaptionskip}{-1pt}
\setlength{\belowcaptionskip}{-15pt}
  \centering
\includegraphics[width=0.9\linewidth]{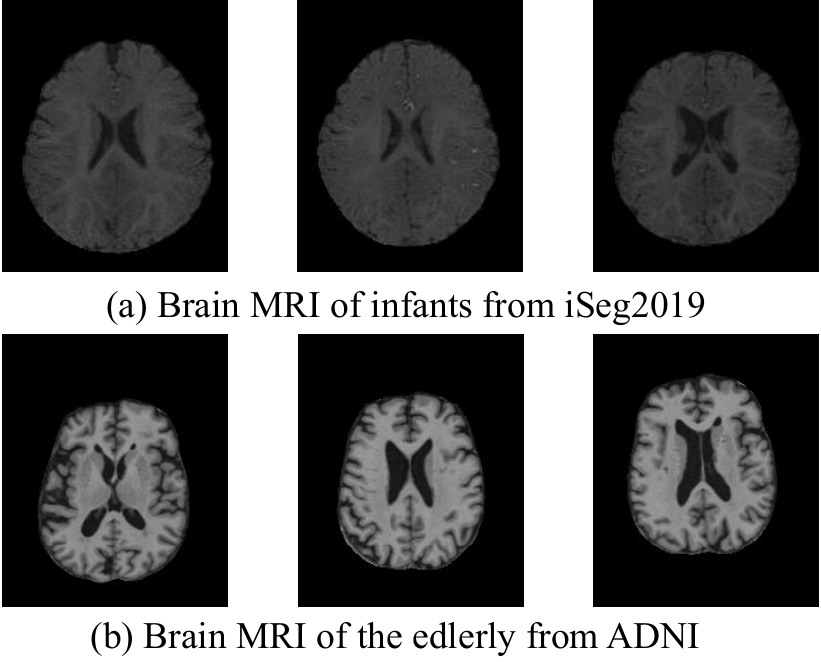}
   \caption{ (a) Brain morphology and tissue contrast of 6-month-old infants from the iSeg2019
dataset. (b) Aging and Alzheimer processes of the elderly from
the ADNI dataset.}
\label{fig:data_example}
\end{figure}

Accurate brain tissue segmentation is essential in diverse neuroscience and clinical studies,  e.g., population analyses of brain cortical architectures and individualized diagnosis of brain diseases.
Due to spatiotemporally dynamic changes in brain structures and functions across the human lifespan, brain magnetic resonance images (MRIs) present longitudinally heterogenous appearances, leading to varying segmentation difficulties at different periods~\cite{9339962}.
Infancy and the elderly are two particularly challenging times for tissue segmentation, as a result of (atypical) developmental and degenerative processes.
For example, as shown in Fig.~\ref{fig:data_example}~(a), the brains of infants before six month old have inverse contrasts between gray matter (GM) and white matter (WM) compared with following periods, and the isointense phase around six-month-old exhibits extremely low inter-tissue contrast.
In contrast, the brains of the elderly and patients with Alzheimer's disease have enlarged cerebrospinal fluid (CSF) and atrophied GM, as shown in Fig.~\ref{fig:data_example}~(b).
These challenges significantly hamper the generalization and longitudinal consistency of existing learning-based methods for automatic brain tissue segmentation~\cite{milletari2016v, WU2022103541, lee2023fine, fan2022attention}, especially considering that labeled training samples for specific time points could be very limited in practice.

To address the above challenges, some methods in the literature~\cite{ren2022local, ouyang2021self} designed contrastive self-supervised learning (SSL) strategies to learn longitudinally consistent representations, which are then fine-tuned on a specific age group for downstream tasks like tissue segmentation.
Such SSL methods commonly have three technical limitations from the application perspective.
That is, they usually require longitudinally paired images of the same subjects in the pre-training phase, which is practically hard to satisfy.
Besides, the fine-tuned segmentation models are typically restricted to match up with the appearance of a specific age group, sacrificing/losing the generalization capacity to other groups.
Moreover, fine-tuning a segmentation model is itself a challenging task, especially when there is not enough labeled training data from the target age group.

In this paper, we revisit and reformulate the idea of longitudinally generalized (or age-agnostic) representation learning.
By nature, it is reasonable to make two fundamental assumptions: \textbf{1)} independent of changing MRI appearances, the high-order differences between different tissue types (e.g., in semantic space) are relatively stable in terms of the time trajectory; \textbf{2)} accurate tissue segmentation requires a reliable mapping function to seamlessly fuse semantic information with age-specific high-resolution image details.
In line with such assumptions, we propose a unified meta-learning framework to concurrently learns to learn a universal feature extractor (e.g., the encoder) for age-agnostic (i.e., longitudinally consistent) anatomical representation learning and a well-initialized segmentation head (e.g., the decoder) that can be flexibly adapted by few age-specific samples (e.g., one labeled MRI) to establish accurate segmentation models generalizable across the lifespan.

Overall, the technical and practical contributions of this paper are four-fold:
\begin{itemize}
    \item We propose a dual meta-learning (DuMeta) paradigm for the construction of longitudinally generalized segmentation networks. Our DuMeta unifies the advantages of both meta-feature learning and meta-initialization learning~\cite{liu2021investigating, hospedales2021meta} in a compact bi-level optimization framework, which contributes to the joint learning of an age-agnostic plug-and-play feature extractor and a reliably pre-trained segmentation head that can be efficiently adapted to different age groups.
    \item For the purpose of learning to learn a universal feature extractor generalizable across the lifespan, we design an intra-tissue temporal similarity regularization and an inter-tissue spatial orthogonality regularization, which are combined together to encourage longitudinal consistency in hierarchically multi-scale representation learning.
        In contrast to previous SSL works, our design explicitly considers class information and has no need of longitudinally paired training data.
    \item Our DuMeta coupled with the two class-aware regularization terms features a practically attractive meta-learning strategy.
     It only needs cross-sectional training samples in the meta-learning stage and as less as one labeled image from an unseen age group to establish an accurate brain tissue segmentation model.
    \item Under the challenging experimental setting of one-shot segmentation, our method significantly outperformed the state-of-the-art longitudinally consistent learning methods on both the infant and elderly datasets.
\end{itemize}

\section{Related works}\label{sec:related-work}

\subsection{Meta-Feature Learning}
Meta-learning typically consists of two fundamental stages, i.e., meta-training and meta-test~\cite{hospedales2021meta}.
As a representative type of meta-learning methods, meta-feature learning (MFL)~\cite{liu2021investigating} aims to learn a common feature extractor in the meta-training stage to share across various (correlated) tasks, based on which a task-specific head is further learned in the meta-test stage for downstream (unseen) tasks.
The common feature extractor can be formulated as an explicit meta-learner, usually meta-trained on various tasks/datasets in a bi-level optimization fashion.
In terms of the strategies to compute the gradients (\emph{w.r.t.} meta-learner parameters), MFL is typically split into two categories: one updates the meta-learner with gradients in the implicit function relation by performing several steps of gradient descent on the loss function~\cite{foo2007efficient, chapelle2002choosing, seeger2008cross, calatroni2017bilevel, lorraine2018stochastic}, the other derives the hyper-gradients through the implicit function theory~\cite{domke2012generic, franceschi2018bilevel, okuno2018ell_p, franceschi2017bridge, baydin2014automatic}. The former is widely used to solve the bi-level optimization problem by automatic differentiation.
For example, Franceschi~\etal~\cite{franceschi2018bilevel} proposed to treat the final layer and remaining layers of a classification network as the base-learner and meta-learner, respectively. The learnable weights of the meta-learner are iteratively updated by gradient descent in the back-propagation fashion.
In this paper, we separate a segmentation network as a plug-and-play feature extractor (i.e., the meta-learner) and a task-specific segmentation head (i.e., the base-learner).
We meta-train the plug-and-play meta-learner by gradient descent in the implicit function relation for longitudinally consistent learning of anatomical representations, under the constraints by two dedicated class-aware regularizations.

\begin{figure*}
\setlength{\abovecaptionskip}{-1pt}
\setlength{\belowcaptionskip}{-15pt}
  \centering
   \includegraphics[width=0.9\linewidth]{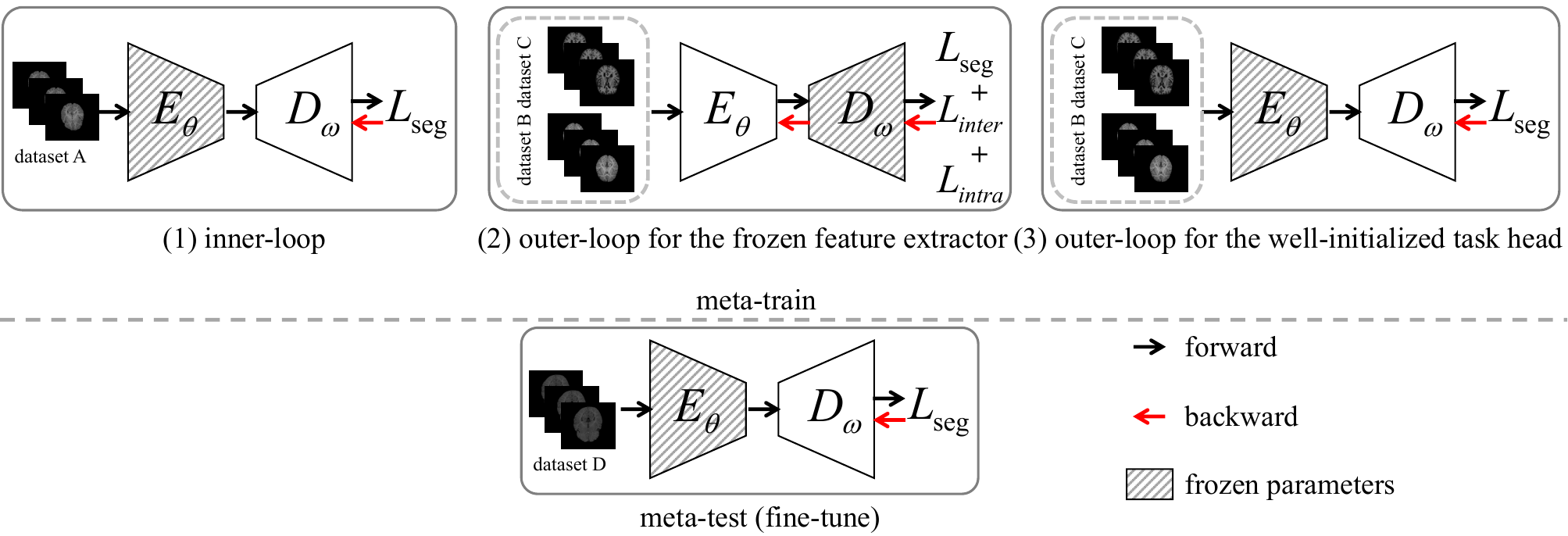}
   \caption{Overview of DuMeta. The meta-train of our DuMeta is divided into three steps (one shared inner-loop and two different outer-loops) with a feature extractor and a task head. Steps (1) and (2) constitute MFL for the frozen feature extractor; Steps (1) and (3) constitute MIL for the well-initialized task head. We finally only fine-tune the task head on an unseen dataset.}
\label{fig:method}
\end{figure*}

\subsection{Meta-Initialization Learning}
As another representative type of meta-learning, meta-initialization learning (MIL) aims to learn a good initialization of network parameters shared across (correlated) tasks \cite{liu2021investigating}.
In contrast to most MFL approaches, MIL typically does not have an explicit meta-learner, while the parameters initialized by MIL can be regarded as an implicit meta-learner.
In addition, the implicit meta-learner of MIL is only explicitly related to the base-learner in the initial state.
A mainstream of MIL methods is to update the initialization parameters with the implicit gradients in the outer-loop of the bi-level optimization framework, with the implicit function relationships coming from the inner-loop.
As a pioneering MIL research, MAML~\cite{finn2017model} uses the implicit gradient of the initialization process to update the meta-learner.
Considering that calculating the hessian matrix in implicit gradient descent is computationally heavy, the following works of MAML have actively investigated the simplification of the MIL steps, e.g., finding ways to directly omits the expensive second-order derivatives~ \cite{li2017meta, nichol2018first, song2019maml}.
For example, ANIL \cite{raghu2019rapid} proposed to remove the inner loop for all but the (task-specific) head of the underlying neural network.
Reptile \cite{nichol2018first} proposed a variety of first-order algorithms to avoid the consumption of the second-order derivatives.
As for the applications of MIL, some methods proposed to use MIL in domain generalization tasks~\cite{liu2020shape, liu2021semi, lee2021meta}.
For example, SAML~\cite{liu2020shape} explicitly simulates domain shifts with virtual meta-train and meta-test during the training procedure to improve the generalization of learned networks.
Notably, most of these existing MIL strategies were formulated to learn the entire network initialization, which by nature is a practically challenging task.
In contrast, our DuMeta simplifies the task: Based on the learning of a frozen feature extractor for longitudinally consistent representation learning, we only need to further learn a lightweight segmentation head with reliable initializations. To this end, we simulate domain shift during fine-tuning by MIL.

\subsection{Longitudinally Consistent Learning}

Self-supervised learning (SSL) is the most actively studied strategy to learn longitudinally consistent or generic representations from data acquired at different time points~\cite{chen2019self,chaitanya2020contrastive, zeng2021positional, park2020contrastive}.
Such SSL methods typically design contrastive regularization terms to learn on longitudinal data the initialized parameters of a deep network, which will then be fine-tuned for downstream tasks.
For example, as an extension of SimSiam~\cite{chen2021exploring}, Ren~\etal~\cite{ren2022local} proposed a patch-wise spatiotemporal similarity loss to encourage longitudinal consistency, and leveraged orthogonality, variance, and covariance regularizations to avoid SSL mode collapse.
Ouyang~\etal proposed an LNE method~\cite{ouyang2021self} that adopts longitudinal neighbourhood embedding to promote the low-dimensional representation to be instructive while keeping a smooth progression trajectory field in the latent space.
For lifespan segmentation, Chen~\etal~\cite{chen2023brain} proposed a joint feature regularization to encourage longitudinal consistency.
Compared with these SSL methods, we take the goal of longitudinally consistent representation learning further.
Specifically, our DuMeta jointly learns a plug-and-play feature extractor with promising longitudinal consistency and a segmentation head with reliable initializations. Only the lightweight segmentation head needs to be fine-tuned for specific age groups. Also, the meta-training of the shared feature extractor does not require longitudinally paired training data.
Moreover, instead of using existing patch- and image-based~\cite{ouyang2021self,ren2022local} regularizations, we propose two class-aware regularizations dedicated for brain tissue segmentation.

\section{Method} \label{sec:method}

\subsection{DuMeta Paradigm}

We propose a dual meta-learning (DuMeta) paradigm for accurate brain tissue segmentation across the lifespan, with the schematic diagram shown in Fig.~\ref{fig:method}.
In the meta-learning (or pre-training) stage, our DuMeta attempts to learn concurrently a universal plug-and-play feature extractor (i.e., $E_{\theta}(\cdot)$) for longitudinally consistent learning of anatomical representations and a well-initialized segmentation head (i.e., $D_{\omega}(\cdot)$).
They are formulated as two different outer-loops under a unified inner-loop in the bi-level meta-learning optimization framework.
In the meta-test (or fine-tuning) stage, only the segmentation head needs to be fine-tuned with few labeled images to efficiently adapt to specific unseen age groups.

\subsubsection{Meta-Learning of Universal Feature Extractor} \label{subsubsec:mfl}
To learn a longitudinally consistent (i.e., age-agnostic) $E_{\theta}(\cdot)$ that can be plug-and-play for unseen age groups, we regard it as an explicit meta-learner and treat $D_{\omega}(\cdot)$ as the base-learner.
Such a universal $E_{\theta}(\cdot)$ is therefore learned in the MFL fashion by leveraging the idea of implicit gradient function.
Specifically, as shown in Fig. \ref{fig:method} (1), in an episode of meta-train, we first freeze $E_{\theta}(\cdot)$ to update $D_{\omega}(\cdot)$ in the inner loop.
To this end, we randomly sample a dataset from the meta-training dataset pool, on which the gradient descent to optimize $D_{\omega}(\cdot)$ is defined as:
{\setlength\abovedisplayskip{0.1cm}
\setlength\belowdisplayskip{0.1cm}
\begin{equation}
    \omega^*_t(\theta_{t-1}) = \omega_{t-1} - \alpha *\frac{\partial L_{\text{inner}}(\omega_{t-1}, \theta_{t-1})}{\partial \omega_{t-1}},
\label{eq:1}
\end{equation}}
where $L_{\text{inner}}$ is the loss function (Dice + CE) of the inner-loop; $\theta_{t-1}$ and $\omega_{t-1}$ denote the parameters of the meta-learner and base-learner at the $(t-1)$th iteration, respectively; and $\alpha$ is the learning rate. It is worth noting that $L_{\text{inner}}$ is actually a function of $\theta_{t-1}$. Thus, the parameter $\omega^*_t(\theta_{t-1})$, updated with the gradient generated by $L_{\text{inner}}$, is still a function of $\theta_{t-1}$. Therefore, Eq.~(\ref{eq:1}) can form an implicit function relationship to assist the updating of $E_{\theta}(\cdot)$ in the subsequent outer-loop.

More specifically, considering that the inner-loop can be regarded as a simulated step to fine-tune the base-learner (i.e., the segmentation head) in our task, we hope that the meta-learner (i.e., the feature extractor) can be good enough to produce promising segmentation results when combined with such a fine-tuned base-learner. For this purpose, as shown in Fig.~\ref{fig:method} (2), we further sample another two datasets from the meta-training pool to update $E_{\theta}(\cdot)$.
Thus, the gradient generated by Eq.~(\ref{eq:1}) in updating $D_{\omega}(\cdot)$ is merged to update the universal meta-learner, such as:
{\setlength\abovedisplayskip{0.1cm}
\setlength\belowdisplayskip{0.1cm}
\begin{equation}
\label{eq:2}
    \theta_t = \theta_{t-1} - \alpha * \frac{\partial L_{\text{outer}}}{\partial\theta_{t-1}};
\end{equation}
\begin{align}
\label{eq:3}
    \frac{\partial L_{\text{outer}}}{\partial\theta_{t-1}} = &\frac{\partial L_{\text{outer}}(\theta_{t-1}, \omega^*_t(\theta_{t-1}))}{\partial\theta_{t-1}} \nonumber\\
    + & \frac{\partial L_{\text{outer}}(\theta_{t-1}, \omega^*_t(\theta_{t-1}))}{\partial\omega^*_t} \frac{\partial\omega^*_t}{\partial\theta_{t-1}},
\end{align}}
where $L_{\text{outer}}$ is the combination of the segmentation loss (i.e., Dice + CE) and class-aware regularizations (will be introduced in Sec.~\ref{subsec:regularization}) for training the universal meta-learner in such an outer-loop.
According to Eq.~(\ref{eq:3}), we can see that the outer-loop gradient to determine the updated $\theta_t$ is actually a combination of two parts, i.e., the direct gradient and the indirect gradient, which jointly make sure the optimization of $E_{\theta}(\cdot)$ working in the right direction to deduce a universal meta-learner generalizable across datasets from different age groups.
That is, the direct gradient part is determined by minimizing $L_{\text{outer}}$, which allows the updated $E_{\theta}(\cdot)$ to perform well in the current outer-loop. On the other hand, the indirect gradient part comes from the implicit function formulated by Eq.~(\ref{eq:1}), which allows the updated $E_{\theta}(\cdot)$ to match up with the updated $D_{\omega}(\cdot)$ to perform well for the simulated fine-tuning step and therefore unseen age groups.
In more detail, the indirect gradient part of Eq.~(\ref{eq:3}) can be unrolled as explicit second-order derivatives:
\begin{align}
\label{eq:4}
    \frac{\partial\omega^*_t}{\partial\theta_{t-1}} = & \frac{\partial(\omega_{t-1} - \alpha *\frac{\partial L_{\text{inner}}(\omega_{t-1}, \theta_{t-1})}{\partial \omega_{t-1}})}{\partial\theta_{t-1}} \nonumber\\
    = & \alpha *\frac{\partial^2 L_{\text{inner}}(\omega_{t-1}, \theta_{t-1})}{\partial\omega_{t-1}\partial\theta_{t-1}}.
\end{align}
 Finally, be substituting Eqs.~(\ref{eq:3}) and~(\ref{eq:4}) into Eq.~(\ref{eq:2}), we get the detailed gradient-descent strategy to optimize $E_{\theta}(\cdot)$:
{\setlength\abovedisplayskip{0.1cm}
\setlength\belowdisplayskip{0.1cm}
\begin{align}
\label{eq:5}
    \theta_t = & \theta_{t-1} - \alpha*\frac{\partial L_{\text{outer}}(\theta_{t-1}, \omega^*_t(\theta_{t-1}))}{\partial\theta_{t-1}}\nonumber\\
    - & \alpha^2*\frac{\partial L_{\text{outer}}(\theta_{t-1}, \omega^*_t(\theta_{t-1}))}{\partial\omega^*_t}\frac{\partial^2 L_{\text{inner}}(\omega_{t-1}, \theta_{t-1})}{\partial\omega_{t-1}\partial\theta_{t-1}},
\end{align}}
which consists of a direct first-order gradient for the current outer-loop and an indirect second-order gradient for the simulated fine-tuning step.

\subsubsection{Meta-Learning of  Initialized Segmentation Head} \label{subsubsec:mil}
Given the plug-and-play feature extractor $E_{\theta}(\cdot)$ determined by Sec.~\ref{subsubsec:mfl}, we further learn a well-initialized segmentation head $D_{\omega}(\cdot)$ in the MIL fashion.
In this step, $E_{\theta}(\cdot)$ is fixed, while the initial weights of $D_{\omega}(\cdot)$, say $\phi_{t-1}$, can be regarded as an implicit meta-learner required to be updated.

To this end, we first assign the initial weight $\phi_{t-1}$ to the segmentation head parameter $\omega$:
{\setlength\abovedisplayskip{0.1cm}
\setlength\belowdisplayskip{0.1cm}
\begin{equation}
    \omega_{t-1} = \phi_{t-1}.
\end{equation}}
Then, we randomly sample a dataset from the meta-training pool to update $\omega_{t-1}$.
In line with the settings of MAML, here we only update $\omega_{t-1}$ by one inner-loop iteration, i..e,
{\setlength\abovedisplayskip{0.1cm}
\setlength\belowdisplayskip{0.1cm}
\begin{equation} \label{equ:inner-mil}
    \omega^*_t = \omega_{t-1} - \alpha *\frac{\partial L_{\text{inner}}}{\partial \omega_{t-1}},
\end{equation}}
It is worth noting that, by the nature of network pre-training, the parameters of the base-learner can be seen as a function (identity function) of the meta-learner parameters.
Hence, after the inner-loop optimization and given the fixed $E_{\theta}(\cdot)$, we update the initialization parameters of the segmentation head (i.e., the implicit meta-learner) in the subsequent outer-loop.
Specifically, we use the updated $\omega^*_t$ to calculate the loss function on another two datasets from the meta-learning pool and update the initialization parameter $\phi_{t-1}$ with gradient descent, such as
\begin{equation}
\label{eq:8}
    \phi_t = \phi_{t-1} - \alpha * \frac{\partial L_{\text{outer}2}}{\partial \phi_{t-1}};
\end{equation}
\begin{equation}
\label{eq:9}
    \frac{\partial L_{\text{outer}2}}{\partial \phi_{t-1}} = \frac{\partial L_{\text{outer}2}}{\partial\omega^*_t}\frac{\partial\omega^*_t}{\partial \phi_{t-1}};
\end{equation}
\begin{align}
\label{eq:10}
    \frac{\partial\omega^*_t}{\partial \phi_{t-1}} = & \frac{\partial(\omega_{t-1} - \alpha *\frac{\partial L_{\text{inner}}}{\partial \omega_{t-1}})}{\partial \phi_{t-1}}\nonumber\\
    = & 1 - \alpha * \frac{\partial^2 L_{\text{inner}}}{\partial\phi_{t-1}^2},
\end{align}
where $L_{\text{outer}2}$ is the same (Dice + CE) segmentation loss as $L_{\text{inner}}$.
By substituting Eqs.~(\ref{eq:9}) and~(\ref{eq:10}) into Eq.~(\ref{eq:8}), the final gradient descent in the current outer-loop iteration is:
{\setlength\abovedisplayskip{0.1cm}
\setlength\belowdisplayskip{0.1cm}
\begin{equation}
    \phi_t = \phi_{t-1} - \alpha * \frac{\partial L_{\text{outer}2}}{\partial \omega_t^*}(1 - \alpha * \frac{\partial^2 L_{\text{inner}}}{\partial\phi_{t-1}^2}).
\label{eq:11}
\end{equation}}
Considering that $\alpha * \frac{\partial^2 L_{\text{inner}}}{\partial\phi_{t-1}^2}$ significantly increases computational complexity and may cause gradient explosion, such a high-level differentiation can be omitted in practice. Besides, Reptile~\cite{nichol2018first} points out that the first-order term can still enhance generalization by increasing the inner product between gradients of different minibatches.

By summarizing the inner-loop and outer-loop iterations in both the MFL step (i.e., Sec.~\ref{subsubsec:mfl}) and the MIL step (i.e., Sec.~\ref{subsubsec:mil}), we can see that these two meta-learning steps actually have the same format of the inner-loop operation (i.e., Eqs.~(\ref{eq:1}) and ~(\ref{equ:inner-mil})).
Therefore, the MFL of a universal feature extractor and the MIL of a well-initialized segmentation head are highly correlated and can be merged to share one single inner-loop for simplicity, resulting in a compact dual meta-learning (DuMeta) paradigm in Algorithm~\ref{alg1}.

\begin{algorithm}[t]
	\caption{Dual Meta-Learning (DuMeta)}
	\label{alg1}
	\begin{algorithmic}[1]
		\REQUIRE Meta-learning pool (of three datasets: A, B, and C); feature extractor $E_{\theta}(\cdot)$ parameterized by $\theta$; segmentation head $D_{\omega}(\cdot)$ parameterized by $\omega$.
  \STATE Randomly initialize $\theta$ and $\omega$;
		\WHILE{Not converged}
		\STATE \textbf{The inner loop:}
            \STATE Sample a mini-batch from a random dataset;
		\STATE Update ${\omega}$ based on Eq.~(\ref{eq:1});	
		\STATE \textbf{The outer-loop for the frozen feature extractor:}
  \STATE Sample a mini-batch from the other two datasets;
		\STATE Update ${\theta}$ based on Eq.~(\ref{eq:5});	
  		\STATE \textbf{The outer-loop for the well-initialized task head:}
      \STATE Sample a mini-batch from the other two datasets;
      \STATE Update ${\omega}$ based on Eq.~(\ref{eq:11}).
		\ENDWHILE
	\end{algorithmic}
\end{algorithm}

\subsection{Class-Aware Regularizations} \label{subsec:regularization}
In Sec.~\ref{subsubsec:mfl}, our DuMeta adopts $L_{outer}$ that combines general segmentation losses with two class-aware regularizations to meta-learn the plug-and-play feature extractor $E_{\theta}(\cdot )$ for age-agnostic (i.e., longitudinally consistent) learning of anatomical representations.
These two class-aware regularizations are intra-tissue temporal similarity and inter-tissue spatial orthogonality, which are designed under a fundamental assumption that, independent of changing time points and MRI appearances, the high-order differences between different tissue types are relatively stable across the lifespan.
Different from existing SSL approaches~\cite{ren2022local, ouyang2021self} that typically implement their contrastive losses to align random MRI patches or images regardless of class information, our class-aware regularizations can be seen as a supervised contrastive learning design to align directly the anatomical presentations of each tissue type, which could be more straightforward for the downstream segmentation task.
Besides, we implement such regularizations at multiple learning scales of $E_{\theta}(\cdot )$ to encourage hierarchically multi-scale longitudinal consistency across age groups.

Specifically, assume that $\{\mathbf{F}_k\}_{k=1}^K$ is a set of the feature maps at $K$ different scales of $E_{\theta}(\cdot )$.
We have $\mathbf{F}_k \in \mathbb{R}^{BS \times NC_k \times H_k \times W_k \times D_k}$, where $BS$, $NC_k$, and $H_k \times W_k \times D_k$ stand for mini-batch size, channel size, and spatial size, respectively.
For each $\mathbf{F}_k$, we first resize the (pseudo) tissue maps (say $\mathbf{M}$) to its spatial resolution, denoted as $\mathbf{M}_k$.
Then, according to the label information in $\mathbf{M}_k$, we get the spatial indices of the voxels belonging to the same classes, based on which we average the corresponding voxel-wise representations from $\mathbf{F}_k$ to obtain the anatomical representations of different tissue types (i.e., GM, WM, and CSF), denoted as $\mathbf{f}_k^{\text{GM}}$, $\mathbf{f}_k^{\text{WM}}$, and $\mathbf{f}_k^{\text{CSF}}$ ($\in \mathbb{R}^{BS \times NC_k}$), respectively.

In terms of $\mathbf{f}_k^{\text{GM}}$, $\mathbf{f}_k^{\text{WM}}$, and $\mathbf{f}_k^{\text{CSF}}$, we define the inter-tissue spatial orthogonality loss as
{\setlength\abovedisplayskip{0cm}
\setlength\belowdisplayskip{0cm}
\begin{align}
    L_{\text{inter}} = & \frac{1}{3K}\sum_{k=1}^K\left(cossim(\mathbf{f}_k^{\text{CSF}}, \mathbf{f}_k^{\text{GM}}) +
    cossim(\mathbf{f}_k^{\text{CSF}}, \mathbf{f}_k^{\text{WM}})\right.\nonumber\\
    + & \left.cossim(\mathbf{f}_k^{\text{GM}}, \mathbf{f}_k^{\text{WM}})\right),
    \label{eq:12}
\end{align}}
where $cossim(\cdot)$ denotes the cosine similarity that we use to encourage the anatomical representations of different tissue types to be separated for each spatial scale.

On the other hand, to promote the anatomical representations of the same tissue at different time points to be consistent, we define the intra-tissue temporal similarity loss as
\begin{align}
    L_{\text{intra}} = &-\frac{1}{3K}\sum_{k=1}^K\left(cossim(\mathbf{f}_{B_k}^{\text{CSF}}, \mathbf{f}_{NC_k}^{\text{CSF}}) \right. \nonumber\\
    + & \left. cossim(\mathbf{f}_{B_k}^{\text{GM}}, \mathbf{f}_{NC_k}^{\text{GM}}) +cossim(\mathbf{f}_{B_k}^{\text{WM}}, \mathbf{f}_{NC_k}^{\text{WM}})\right),
    \label{eq:13}
\end{align}
where $\mathbf{f}_{B_k}^{\text{CSF}}$ and $\mathbf{f}_{NC_k}^{\text{CSF}}$ denote the $k$th-scale tissue anatomical representations quantified on two datasets (from different age groups), respectively.

Finally, the overall loss function in the outer-loop of our MFL stage to update the universal feature extractor $E_{\theta}(\cdot )$ can be specified as:
\begin{equation}
    L_{\text{outer}} = L_{\text{seg}} + \beta L_{\text{inter}} + \gamma L_{\text{intra}},
\end{equation}
where $\beta$ and $\gamma$ are two hyper-parameters, which were set as 0.1 and 0.001, respectively, in our experiments.

\section{Experiments}
\label{sec:results}

\subsection{Datasets}
To evaluate the efficacy of our method, we applied it to brain tissue segmentation on two difficult age groups, i.e., 6-month-old infancy and the elderly, under the challenging setting of one-shot segmentation.
Specifically, we meta-trained the segmentation network (consisting of a feature extractor and a segmentation head) on three public datasets, including OASIS3, IBIS12M, and IBIS24M, and meta-tested it on another two public datasets, i.e., ADNI and iSeg-2019.
The network inputs were T1w MRIs undergone a series of pre-processing, including skull stripping, bias field correction, and intensity normalization.
To obtain high-quality meta-training labels, we segmented the pre-processed T1w images automatically by using an advanced pipeline, i.e., iBEAT~\cite{wang2023ibeat}. These tissue maps produced by iBEAT can be regarded as pseudo labels from the aspect of semi-supervision.
To obtain accurate meta-test labels, iBEAT was first applied, followed by manual corrections by experienced neuroradiologists to produce the ground-truth tissue maps.
Each of the three meta-training datasets was split according to an 80-20 training-validation ratio, where the validation set was used for model and hyper-parameter selection.
After meta-training, one training subject from each of the two meta-test datasets was randomly sampled to fine-tune the segmentation head. We reported the metrics on the test set from the meta-test datasets.

\paragraph{OASIS3~\cite{lamontagne2019oasis}:}
A publicly accessible dataset consisting of 1,639 T1w MRIs from 992 longitudinally scanned participants (each with 1--5 temporal acquisitions across a 5-year observation window). The participants include Alzheimer's disease (AD) patients, and healthy controls or minimally impaired abilities, with the ages ranging from 42 to 95.

\paragraph{ADNI~\cite{mueller2005Alzheimer}:}
A multisite dataset consisting of 2,389 longitudinal T1w MRIs (at least two visits per subject). Participants with AD, those who may acquire AD, and healthy controls without any evidence of cognitive impairment were included, with the ages ranging from 20 to 90.

\paragraph{IBIS:}
An infant brain imaging dataset that collects 1,272 structural T1w/T2w MRIs from 552 babies throughout 3 to 36 months old, including both controls and infants at high risk for Autism Spectrum Disorder (ASD).
These subjects can be assigned into different groups according to their ages. For example, IBIS12M and IBIS24M denote the 12-month-old and 24-month-old groups, respectively.

\paragraph{iSeg-2019~\cite{9339962}:}
The benchmark of the MICCAI 2019 grand challenge on 6-month infant brain MRI segmentation from multiple sites~\cite{9339962}. The iSeg-2019 training and test sets include 10 and 13 subjects, respectively.
Each subject consists of T1w and T2w MRIs. Notably, 6-month-old infant brain segmentation is challenging, considering that the GM and WM intensity ranges are substantially overlapped (particularly around the cortical areas) at this isointense stage, which results in the ambiguities that present the greatest obstacle for tissue segmentation.

\begin{table*}[ht]
\small
\centering
\caption{One-shot segmentation on the official validation set of the iSeg-2019 dataset.}
\label{tab:comp}
\begin{tabular}{l|ccc|ccc}
\hline
\multirow{2}{*}{Exp}           & \multicolumn{3}{c|}{Dice↑}                       & \multicolumn{3}{c}{ASD↓}                         \\ \cline{2-7}
                               & CSF            & GM             & WM             & CSF            & GM             & WM             \\ \hline
RandInitUnet.2D                & 0.8767±0.0113 & 0.8402±0.0220 & 0.7965±0.0035 & 0.3317±0.0215 & 0.5578±0.0825 & 0.8479±0.1930 \\
RandInitUnet.3D                & 0.9029±0.0066 & 0.8616±0.0176 & 0.8200±0.0071 & 0.2652±0.0162 & 0.5497±0.0658 & 0.6487±0.1324 \\
Context Restore \cite{chen2019self}  & 0.9070±0.0065 & 0.8615±0.0118 & 0.8245±0.0155 & 0.2641±0.0239 & 0.5420±0.0560 & 0.6298±0.1068 \\
LNE \cite{ouyang2021self}            & 0.9315±0.0115 & 0.8937±0.0147 & 0.8654±0.0141 & 0.1753±0.0163 & 0.4482±0.0373 & 0.5274±0.0710 \\
GLCL \cite{chaitanya2020contrastive} & 0.9141±0.0103 & 0.8718±0.0185 & 0.8337±0.0088 & 0.2337±0.0147 & 0.5017±0.0546 & 0.6596±0.1024 \\
PCL \cite{zeng2021positional}        & 0.9139±0.0089 & 0.8688±0.0148 & 0.8313±0.0133 & 0.2326±0.0130 & 0.5161±0.0484 & 0.6692±0.0943 \\
PatchNCE \cite{park2020contrastive}  & 0.9444±0.0096 & 0.9043±0.0081 & 0.8782±0.0223 & 0.1360±0.0153 & 0.3820±0.0376 & 0.4520±0.0760 \\
MAML \cite{finn2017model} & 0.9433±0.0107                    & 0.9025±0.0100                    &  0.8768±0.0185                     & 0.0174±0.0139                    & 0.3931±0.0493                    & 0.5374±0.0859 \\
Reptile \cite{nichol2018first} & 0.9450±0.0129                    & 0.9047±0.0121                    &  0.8777±0.0143                     & 0.0117±0.0123                    &  0.3896±0.0451                    & 0.5324±0.0825 \\
Ours                           & \textbf{0.9611±0.0059}  & \textbf{0.9313±0.0083}  & \textbf{0.9145±0.0126}  & \textbf{0.1082±0.0158}  & \textbf{0.2916±0.0423}  & \textbf{0.3318±0.0483}  \\ \hline
\end{tabular}
\centering
\caption{One-shot segmentation on the ADNI dataset.}
\label{tab:comp_ADNI}
\begin{tabular}{l|ccc|ccc}
\hline
\multirow{2}{*}{Exp}           & \multicolumn{3}{c|}{Dice↑}                       & \multicolumn{3}{c}{ASD↓}                         \\ \cline{2-7}
                               & CSF            & GM             & WM             & CSF            & GM             & WM             \\ \hline
RandInitUnet.2D                & 0.9223±0.0084 & 0.9051±0.0098 & 0.9361±0.0096 & 0.1571±0.0422 & 0.1573±0.0338 & 0.2059±0.0909 \\
RandInitUnet.3D                & 0.9459±0.0062 & 0.9236±0.0073 & 0.9495±0.0058 & 0.0933±0.0205 & 0.1160±0.0297 & 0.1226±0.0291 \\
Context Restore \cite{chen2019self}  & 0.9500±0.0058 & 0.9289±0.0069 & 0.9527±0.0054 & 0.0821±0.0179 & 0.1056±0.0267 & 0.1101±0.0241 \\
LNE \cite{ouyang2021self}            & 0.9665±0.0043 & 0.9456±0.0065 & 0.9637±0.0047 & 0.0487±0.0107 & 0.0711±0.0168 & 0.0760±0.0180 \\
GLCL \cite{chaitanya2020contrastive} & 0.9531±0.0055 & 0.9332±0.0069 & 0.9559±0.0052 & 0.0736±0.0151 & 0.0964±0.0238 & 0.1018±0.0236 \\
PCL \cite{zeng2021positional}        & 0.9526±0.0057 & 0.9314±0.0068 & 0.9547±0.0053 & 0.0750±0.0137 & 0.0995±0.0255 & 0.1059±0.0273 \\
PatchNCE \cite{park2020contrastive}  & 0.9716±0.0040 & 0.9520±0.0060 & 0.9687±0.0043 & 0.0402±0.0091 & 0.0591±0.0118 & 0.0634±0.0157 \\ 
MAML \cite{finn2017model} & 0.9680±0.0062                    & 0.9472±0.0070                    & 0.9647±0.0061                     & 0.0460±0.0197                    & 0.0680±0.0239                    & 0.0736±0.0281 \\
Reptile \cite{nichol2018first} & 0.9687±0.0056                    & 0.9476±0.0071                    &  0.9656±0.0049                     & 0.0452±0.0167                    & 0.0666±0.0163                    &  0.0702±0.0176 \\
Ours                           & \textbf{0.9809±0.0021} & \textbf{0.9678±0.0034} & \textbf{0.9796±0.0024} & \textbf{0.0222±0.0042} & \textbf{0.0315±0.0052} & \textbf{0.0322±0.0054} \\ \hline
\end{tabular}
\end{table*}

\subsection{Implementation Details}
We used 3D U-Net~\cite{ronneberger2015u} as our baseline model, which performs five down-samplings and five up-samplings in the encoding and decoding path, respectively.
We used InstanceNorm3d for normalization and ReLU for nonlinear activation.
The network was trained by the SGD optimizer with Nesterov momentum ($\mu$ = 0.99), and poly learning rate schedule (initial 0.01). The weight decay was set as 3e-5 to avoid overfitting.
The size of input image patches was $128\times128\times128$, and the mini-batch size was 2.
We conducted deep supervision on five scales, with the weights of 0.0625, 0.125, 0.25, 0.5, and 1 (from coarse to fine), respectively.
The training data was extensively augmented using the pipeline of nnUNet~\cite{isensee2021nnu}.
Two commonly used metrics were selected to quantify segmentation performance: Dice ratio and Average Symmetric Surface Distance (ASD).
A larger dice value indicates better overlapping between the prediction and ground truth.
In contrast, smaller ASD values suggest better performance in terms of the boundary consistency between the prediction and ground truth.

\begin{figure}[t]
\setlength{\abovecaptionskip}{-5pt}
\setlength{\belowcaptionskip}{-20pt}
  \centering
   \includegraphics[width=0.99\linewidth]{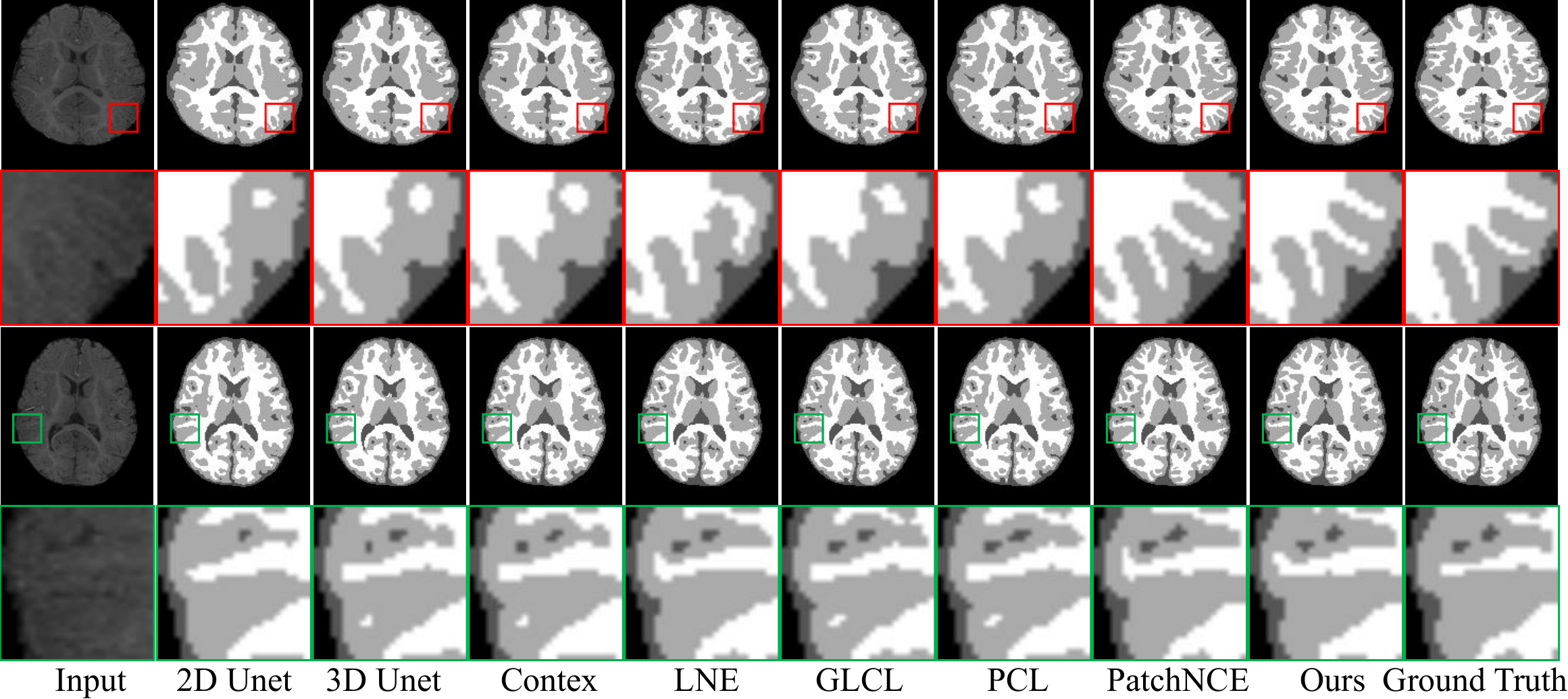}
   \caption{The 2D slice views of representative one-shot segmentation results on the held-out test set of iSeg-2019.}
\label{fig:comp_iSeg_2D}
\setlength{\abovecaptionskip}{-5pt}
\setlength{\belowcaptionskip}{-20pt}
  \centering
   \includegraphics[width=0.99\linewidth]{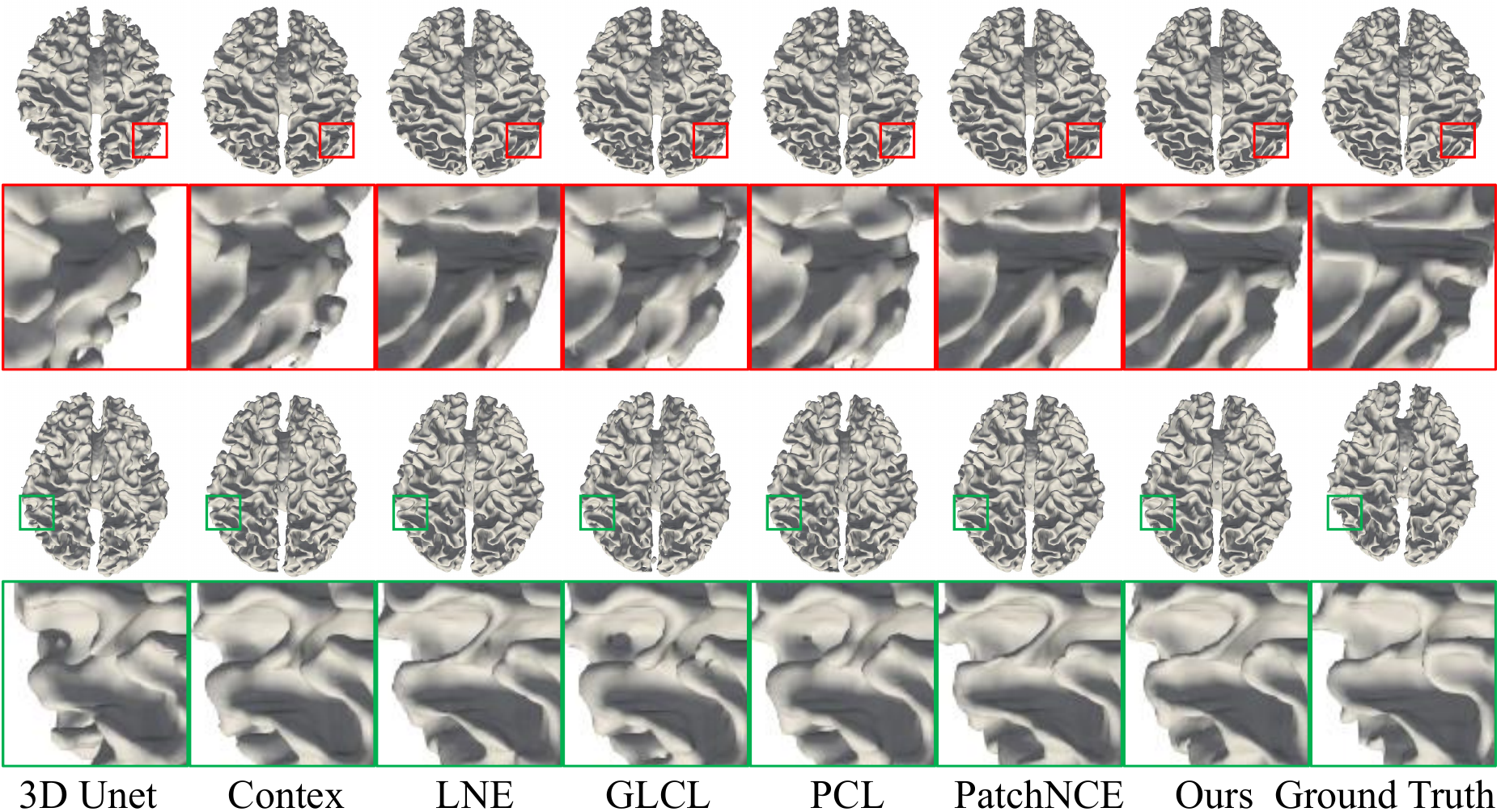}
   \caption{The 3D WM surface views of representative one-shot segmentation results on the held-out test set of iSeg-2019.}
\label{fig:comp_iSeg_3D}
\end{figure}

\begin{figure}[t]
\setlength{\abovecaptionskip}{-1pt}
\setlength{\belowcaptionskip}{-15pt}
  \centering
   \includegraphics[width=0.99\linewidth]{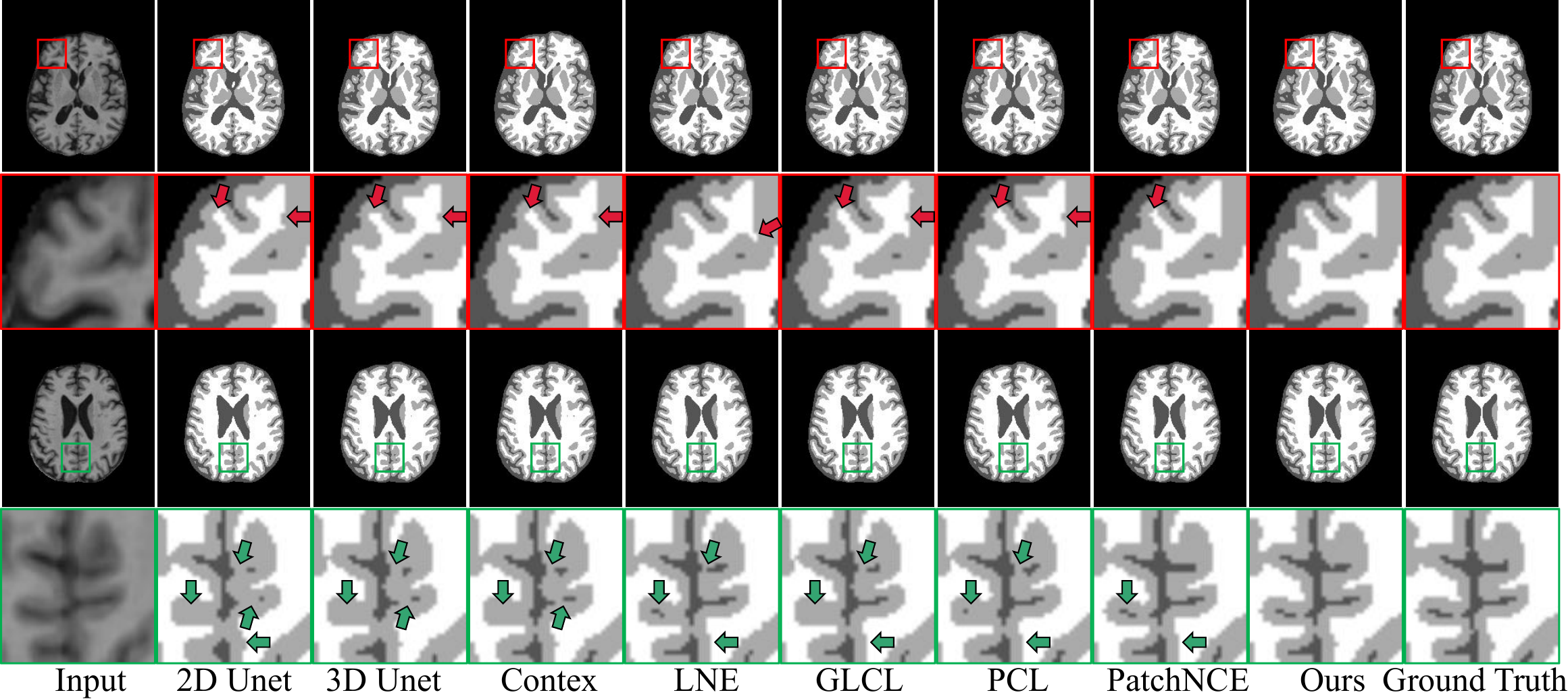}
   \caption{The 2D slice views of representative one-shot segmentation results on the ADNI dataset.}
\label{fig:comp_ADNI_2D}
%
\setlength{\abovecaptionskip}{-1pt}
\setlength{\belowcaptionskip}{-20pt}
  \centering
   \includegraphics[width=0.99\linewidth]{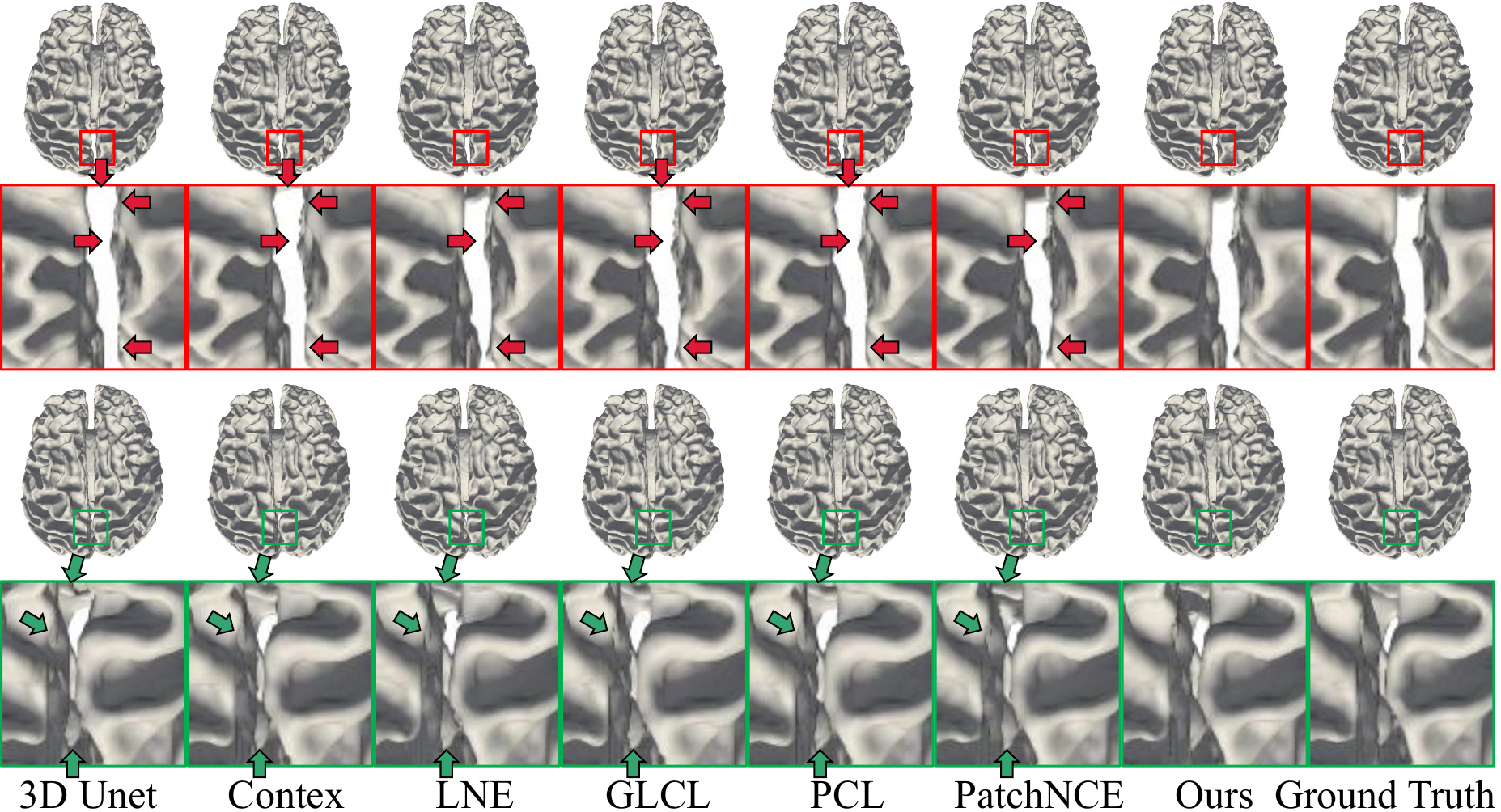}
   \caption{The 3D WM surface views of representative one-shot segmentation results on the ADNI dataset.}
\label{fig:comp_ADNI_3D}
\end{figure}

\subsection{Comparison Results}
We compared our method with other methods addressing longitudinally consistent/generalized representation learning. We also compared our method with other meta-learning methods addressing few-shot learning. In order to verify the performance of our method on longitudinal data, we performed challenging one-shot brain tissue segmentation on the ADNI and iSeg-2019 datasets in the meta-test stage. Both data sets have longitudinal domain shifts due to dynamic neuro-degeneration/development processes.

\paragraph{One-shot segmentation on iSeg-2019:}
The iSeg-2019 dataset consists of 6-month-old infant MRIs from multiple sites, which present a flip GM/WM contrast with a huge domain shift compared with other time points.
Our DuMeta meta-trained (pre-trained) the 3D U-Net on IBIS12M, and IBIS24M (i.e., 12/24-month-old infants), and OASIS3 (i.e., the elderly), and then meta-tested (fine-tuned) it on iSeg-2019.
Notably, our model did not see 6-month-old data during meta-training, and we froze the feature extractor and only fine-tuned the segmentation head with one 6-month-old subject during meta-test.
The quantitative segmentation results obtained by different competing methods are summarized in Table~\ref{tab:comp}. From Table~\ref{tab:comp}, we can see that our DuMeta outperformed other competing methods by a large margin in learning a longitudinally generalizable segmentation model.
As a support to the quantitative evaluations, we also show the qualitative experimental results in Figs.~\ref{fig:comp_iSeg_2D} and~\ref{fig:comp_iSeg_3D}, from which we can see that our method performs better in segmenting detailed structures, e.g., cortical foldings.

\begin{table*}[t]
\setlength{\tabcolsep}{4pt}
\small
\centering
\caption{Ablation study of different components on the held-out test set of iSeg-2019.}
\label{tab:abla1}
\begin{tabular}{lccc|ccc|ccc}
\hline
\multicolumn{1}{c}{\multirow{2}{*}{Exp}} & \multirow{2}{*}{$L_{inter}$} & \multirow{2}{*}{$L_{intra}$} & \multirow{2}{*}{\begin{tabular}[c]{@{}c@{}}DuMeta\end{tabular}} & \multicolumn{3}{c|}{Dice↑}                       & \multicolumn{3}{c}{ASD↓}                         \\ \cline{5-10}
\multicolumn{1}{c}{}                     &                           &                           &                                                                                & CSF            & GM             & WM             & CSF            & GM             & WM               \\ \hline
A                                        &                           &                           &                                     & 0.9155±0.0147 & 0.8724±0.0114 & \multicolumn{1}{c|}{0.8375±0.0094} & 0.1113±0.0194 & 0.2954±0.0498 & 0.4082±0.0955 \\
B                                        &                           &                           & $\checkmark$                                    & 0.9358±0.0060 & 0.8955±0.0011 & \multicolumn{1}{c|}{0.8679±0.0235} & 0.0781±0.0106 & 0.2176±0.0421 & 0.2997±0.0847 \\
C                                        & $\checkmark$                          &                           & $\checkmark$                                    & 0.9431±0.0057 & 0.9015±0.0029 & \multicolumn{1}{c|}{0.8742±0.0247} & 0.0669±0.0093 & 0.2009±0.0428 & 0.2775±0.0826 \\
D                                        &                           & $\checkmark$                          & $\checkmark$                                    & 0.9414±0.0051 & 0.8977±0.0025 & \multicolumn{1}{c|}{0.8690±0.0235} & 0.0701±0.0087 & 0.2127±0.0447 & 0.2935±0.0830 \\
E                                        & $\checkmark$                          & $\checkmark$                          & $\checkmark$                                    & \textbf{0.9492±0.0078} & \textbf{0.9091±0.0001} & \multicolumn{1}{c|}{\textbf{0.8826±0.0205}} & \textbf{0.0597±0.0104} & \textbf{0.1824±0.0328} & \textbf{0.2488±0.0664} \\ \hline
\end{tabular}
\small
\centering
\caption{Ablation study of different base/meta-learner splits on the held-out test set of iSeg-2019 (ft means fine-tune).}
\label{tab:ablation_decoder}
\begin{tabular}{l|ccc|ccc}
\hline
\multirow{2}{*}{Exp} & \multicolumn{3}{c|}{Dice↑}                                                                                    & \multicolumn{3}{c}{ASD↓}                                                                                     \\ \cline{2-7}
                     & CSF                                & GM                                 & WM                                  & CSF                                & GM                                 & WM                                 \\ \hline
w/o ft               & 0.8450±0.0074                     & 0.8082±0.0089                     & 0.7564±0.0091                      & 0.2487±0.0320                     & 0.5298±0.0652                     & 0.7460±0.2265                     \\
ft all               & \multicolumn{1}{l}{0.9454±0.0069} & \multicolumn{1}{l}{0.9076±0.0008} & \multicolumn{1}{l|}{0.8823±0.0220} & \multicolumn{1}{l}{0.0653±0.0076} & \multicolumn{1}{l}{0.1857±0.0369} & \multicolumn{1}{l}{0.2493±0.0715} \\
ft 5 upsample layers & \multicolumn{1}{l}{0.9484±0.0087} & \multicolumn{1}{l}{0.9083±0.0003} & \multicolumn{1}{l|}{0.8820±0.0220} & \multicolumn{1}{l}{0.0617±0.0130} & \multicolumn{1}{l}{0.1842±0.0322} & \multicolumn{1}{l}{\textbf{0.2487±0.0669}} \\
ft 4 upsample layers & 0.9491±0.0081                     & \textbf{0.9096±0.0005}                     & \textbf{0.8829±0.0194}                      & 0.0618±0.0124                     & 0.1841±0.0331                     & 0.2502±0.0644                     \\
ft 3 upsample layers & \textbf{0.9492±0.0078}                     & 0.9091±0.0001                     & 0.8826±0.0205                      & \textbf{0.0597±0.0104}                     & \textbf{0.1824±0.0328}                     & 0.2488±0.0664                     \\
ft 2 upsample layers & 0.9451±0.0076                     & 0.9058±0.0002                     & 0.8793±0.0211                      & 0.0654±0.0101                     & 0.1908±0.0373                     & 0.2600±0.0730                     \\
ft 1 upsample layer  & 0.9450±0.0066                     & 0.9049±0.0013                     & 0.8782±0.0229                      & 0.0661±0.0103                     & 0.1920±0.0394                     & 0.2620±0.0775                     \\
\hline
\end{tabular}
\end{table*}

\paragraph{One-shot segmentation on ADNI:}
By leveraging the same meta-training strategy used in iSeg-2019, we further meta-tested the pre-trained 3D U-Net on the ADNI dataset.
Correspondingly, the quantitative segmentation results and representative visualization results are summarized in Table \ref{tab:comp_ADNI} and Figs.~\ref{fig:comp_ADNI_2D} and~\ref{fig:comp_ADNI_3D}, respectively.
It can be seen that, consistent with the observations on iSeg-2019, our DuMeta also outperformed other competing methods by a large margin in the case of aging brain tissue segmentation.
These results suggest the promising generalization capacity of our DuMeta in segmenting brain tissues across the lifespan.

Besides the state-of-the-art segmentation accuracies, it is worth mentioning that the pre-training by other competing methods was based on self-supervision without ground-truth label information.
Similarly, our DuMeta does not need the ground truth either during meta-training, as we used pseudo labels efficiently generated by iBEAT, which can be understood as semi-supervision.
However, the inclusion of pseudo label information can help avoid mode collapse potentially caused by contrastive self-supervision.
For example, because of the U-Net skip connection, the self-supervised model degenerated into identity mappings can still achieve good self-reconstruction, which hampers the model to learn informative representation for downstream segmentation tasks.
Moreover, in contrast to contrastive self-supervision that typically needs longitudinally paired training, our DuMeta works on unpaired cross-sectional data, which is much easier to satisfy in practice.

\begin{figure}[h]
\begin{center}
   \includegraphics[width=0.8\linewidth]{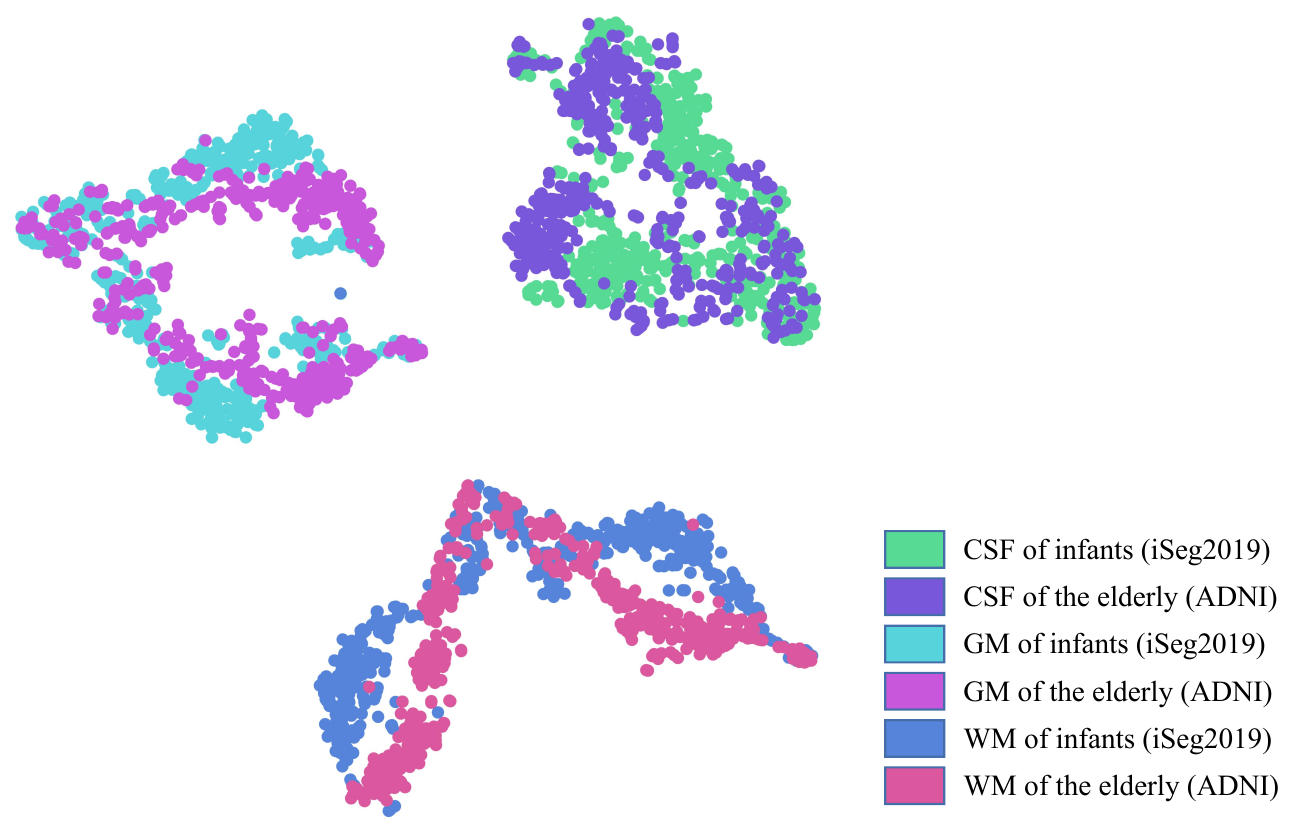}
\end{center}
   \caption{Visualizing meta-learned features for different ages.}
\label{fig:t-sne}
\end{figure}

\paragraph{Evaluation of longitudinal consistency:}
We evaluated the longitudinal consistency of different segmentations on ADNI by using two metrics, i.e.,
STCS \cite{li2021longitudinal} and ASPC
\cite{reuter2012within}. Results in Table
\ref{tab:lc} suggest the better performance of DuMeta. We also conducted a t-sne visualization of
the meta-learned features learned from the ADNI and i-Seg subjects. Examples in Fig. \ref{fig:t-sne} imply that DuMeta can learn reliably time-invariant discriminative representations.

\begin{table}[]
\scriptsize
\centering
\caption{Evaluation of longitudinal consistency on the ADNI dataset.}
\label{tab:lc}
\begin{tabular}{l|ccc|lll}
\hline
\multirow{2}{*}{Exp} & \multicolumn{3}{c|}{STCS↑}                          & \multicolumn{3}{l}{ASPC↓}                                                 \\ \cline{2-7} 
                     & CSF             & GM              & WM              & \multicolumn{1}{c}{CSF} & \multicolumn{1}{c}{GM} & \multicolumn{1}{c}{WM} \\ \hline
RandInitUnet.2D      & 0.8809          & 0.8584          & 0.8998          & 6.35                    & 8.32                   & 5.16                   \\
RandInitUnet.3D      & 0.9063          & 0.8862          & 0.9177          & 5.06                    & 6.02                   & 4.20                   \\
Context Restore \cite{chen2019self}     & 0.8386          & 0.8180          & 0.8580          & 9.15                    & 11.07                  & 8.08                   \\
LNE \cite{ouyang2021self}                 & 0.9015          & 0.8796          & 0.9132          & 4.78                    & 6.81                   & 4.49                   \\
GLCL \cite{chaitanya2020contrastive}                & 0.8956          & 0.8706          & 0.9057          & 5.18                    & 7.57                   & 4.75                   \\
PCL \cite{zeng2021positional}                 & 0.9155          & 0.8968          & 0.9275          & 3.98                    & 5.63                   & 3.13                   \\
PatchNCE \cite{park2020contrastive}            & 0.9250          & 0.9087          & 0.9356          & 3.80                    & 5.05                   & 2.94                   \\
Ours                 & \textbf{0.9373} & \textbf{0.9222} & \textbf{0.9468} & \textbf{2.66}           & \textbf{3.40}          & \textbf{2.20}          \\ \hline
\end{tabular}
\end{table}

\subsection{Ablation Studies}
We performed ablation experiments on the iSeg-2019 training set.
Specifically, iSeg-2019 provides 10 publicly available labeled training samples. In the meta-test stage, we used one sample for fine-tuning, two subjects for validation, and the remaining two subjects for test.

\paragraph{Role of DuMeta and class-aware regularizations:}
To evaluate the efficacy of the DuMeta paradigm and class-aware regularization terms proposed in this paper, we conducted a series of ablation experiments by removing them to check the influence on segmentation accuracies, with the results summarized in Table~\ref{tab:abla1}.
It can be seen that, compared with the baseline (i.e., A in Table~\ref{tab:abla1}), DuMeta led to significantly improved performance, implying its efficacy in learning generalizable segmentation networks across the lifespan.
Also, we can see that both the two class-aware regularizations brought respective performance gains.
Among them, the improvements by the inter-tissue spatial orthogonality are relatively larger than the intra-tissue temporal similarity. The potential reason could be that the inter-tissue spatial orthogonality encourages the differentiation between different class representations and could be more straightforwardly linked to the segmentation task.
The best result was achieved by using both regularizations, which suggest that they are complementary in longitudinally consistent representation learning for segmentation.

\paragraph{Influence of different base/meta-learner splits:}
We have carried out experiments to check the influence brought by different base/meta-learner (i.e., feature extractor/segmentation head) splits of the 3D U-Net, with the results shown in Table~\ref{tab:ablation_decoder}.
It can be seen from Table~\ref{tab:ablation_decoder} that, compared with the pre-trained model (i.e., w/o ft), fine-tuning all parameters (i.e., ft all) achieved much better results, suggesting that there are indeed domain gaps across the lifespan.
On the other hand, when only a small number of parameters were fine-tuned (e.g., ft 1 upsample layer), the performance decreased compared with ft all.
The reason could be that there is a gap between the local and global optimum due to too few parameters available for fine-tuning.
In contrast, when fine-tuning the last three upsample layers (i.e., ft 3 upsample layers) in the meta-test stage, our DuMeta led to the best performance.
These results suggest that there is a trade-off between convergence difficulty and convergence quality in model fine-tuning, considering that fine-tuning the whole network is hard to converge, while fine-tuning few network layers is prone to suboptimal solutions. Our method found a good compromise, leading to accurate and generalizable segmentation results.

\paragraph{Hyperparameter sensitivity evaluation:}
We performed a hyperparameter sensitivity evaluation. Results in Table \ref{tab:abla_hyper} suggest the inter-class term Eq.~(\ref{eq:12}) does have a larger influence than the
intra-class term Eq.~(\ref{eq:13}). This may be due to the fact that the inter-class term has a more direct effect on the segmentation task.
\begin{table}[]
\scriptsize
\centering
\caption{Ablation of hyperparameters on the ADNI dataset.}
\label{tab:abla_hyper}
\begin{tabular}{l|l|ccc|ccc}
\hline
\multirow{2}{*}{beta} & \multirow{2}{*}{gama} & \multicolumn{3}{c|}{Dice↑}                          & \multicolumn{3}{c}{ASD↓}                            \\ \cline{3-8} 
                      &                       & CSF             & GM              & WM              & CSF             & GM              & WM              \\ \hline
1e-1                  & 1e-1                  & 0.9714          & 0.9516          & 0.9684          & 0.0407          & 0.0600          & 0.0647          \\
1e-3                  & 1e-3                  & 0.9710          & 0.9509          & 0.9679          & 0.0417          & 0.0617          & 0.0652          \\
1e-3                  & 1e-1                  & 0.9697          & 0.9490          & 0.9671          & 0.0448          & 0.0656          & 0.0675          \\
1e-1                  & 1e-3                  & \textbf{0.9809} & \textbf{0.9678} & \textbf{0.9796} & \textbf{0.0222} & \textbf{0.0315} & \textbf{0.0322} \\ \hline
\end{tabular}
\end{table}

\section{Conclusion} \label{sec:conclusion}
In this paper, we have proposed a dual meta-learning (DuMeta) paradigm coupled with dedicated class-aware regularizations to learn longitudinally consistent representations from brain MRIs for accurate brain tissue segmentation across the lifespan.
Our DuMeta unifies the advantages of both meta-feature learning and meta-initialization learning to jointly meta-learn an age-agnostic plug-and-play feature extractor and a well-initialized segmentation head.
In the meta-test, only one labeled data is needed by DuMeta to adopt the segmentation head to unseen age groups.
Experiments carried out on the ADNI and iSeg2019 datasets show that our method significantly outperforms existing longitudinally consistent representation learning methods.

\section*{Acknowledgement}
This work was supported in part by STI 2030—Major Projects (No. 2022ZD0209000), NSFC 62101431, and NSFC 62101430.

{\small
\bibliographystyle{ieee_fullname}
\bibliography{egbib}
}

\end{document}